# Fault Area Detection in Leaf Diseases using k-means Clustering

Subhajit Maity[1], Sujan Sarkar[2], Avinaba Tapadar[3], Ayan Dutta[4], Sanket Biswas[5], Sayon Nayek[6], Pritam Saha[7]

[1,2,3,4,6,7]Department of Electronics and Communication Engineering, [5]Department of Computer Science Engineering

Jalpaiguri Government Engineering College

Jalpaiguri, West Bengal, India

smaity.jgec18@gmail.com[1], sujansa19997@gmail.com[2], avinaba.bwn@gmail.com[3], dutta.ayan1998@gmail.com[4], sanketbiswas1995@gmail.com[5], sayon.bwn@gmail.com[6], pritamsaha125@gmail.com[7]

*Abstract*— **With increasing population the crisis of food is getting bigger day by day. In this time of crisis, the leaf disease of crops is the biggest problem in the food industry. In this paper, we have addressed that problem and proposed an efficient method to detect leaf disease. Leaf diseases can be detected from simple images of the leaves with the help of image processing and segmentation. Using k-means clustering and Otsu's method the faulty region in a leaf is detected which helps to determine proper course of action to be taken. Further the ratio of normal and faulty region if calculated would be able to predict if the leaf can be cured at all.**

## I. INTRODUCTION

Based on some research, by 2100 earth's estimated population is 11.2 billion, and with this day to day growing issue, there is an argent need to expand the production of food. As there is there very few number of cultivated lands left, to feed the whole world we have to produce food beyond our limit. It is also observed that crops, of worth several billion dollars are losses annually due to corps diseases. But the main problem is production of food slows down by the influence of diseases. At this moment it will be a necessary step to minimize the loss and to secure the corps by using technological support. In most cases pests and diseases are found on the leaves or branches of the plant.

TABLE I . The Total loss of crop according to Food and Agriculture Organization of the United Nations

| Country (Year) | Estimated Loss (mt) | Loss as Percentage of Expected Output (%) | Value of Production Loss (US$ Million, 1994) |
|---|---|---|---|
| Thailand (1994) | 130 | 58 | 650 |
| Philippines (1989) | 57 | 96 | 285 |
| Ecuador (1992) | 34 | 27 | 170 |
| Indonesia (1991) | 50 | 34 | 250 |
| China (1992) | 180 | 84 | 900 |
| Taiwan (1987) | 100 | 72 | 500 |
| Mexico (1994) | 1 | 8 | 5 |
| USA (1993) | 12 | NA[1] | 60 |
| India (1994) | 25 | 36 | 125 |
| Vietnam (1994) | 10 | 20 | 50 |
| Bangladesh (1994) | 5 | 14 | 25 |
| Total | 541 | 74 | 3,019 |

[1]NA = not available.

Because of the compilation of visual patterns, enough study is not done on these optically observed diseases, and for that, the demand for more precise and sophisticated image pattern discerning is increasing continuously. By using image processing techniques, an image can be defined over two dimensions (feasibly more), using that more precise image pattern can be found, which plays an important role in crops cultivation. There are a number of popular digital image processing techniques are available like Hidden Markov models, Image restoration, Anisotropic diffusion, Image editing, Linear Filtering, Partial differential equations, Independent Component analysis, Pixilation, Principal Components Analysis, Wavelets, Self-organizing.

Classification of crop diseases using image processing was researched by Ying[15]. Ying said, " Leaves with marks must be carefully examined in order to carry out intelligent diagnosis on the basis of image processing".
Important methods of image processing :
- Image clipping: Based on marks , Classification of leaves.
- Thresholding: Image segmentation into spot background.
- Noise reduction: Noises are wiped out by medium filter.

By experts, two different methods for the diagnosing of plant diseases were put forward:
1. Graphical representation
2. Step by step descriptive methods

For the grading process of flue-cured tobacco leaves image feature extraction is useful. In machine vision techniques [16], automated investigation of flue-cured tobacco leaves was mentioned. The above mentioned techniques were used to solve the problems of feature extraction.

## II. LITERATURE SURVEY

In 1979 and 1980 Punjab and Haryana, states of India, heavily infected by a disease Xanthomonas oryzae, a bacterial disease causes most destructive bacterial blight of rice which causes almost 50% of worldwide annual yield loss[21,22].the bacterial blight is most common disease on hybrid rice in Zhejiang Province in China[23] reported by Cai and Zhong.

Another disease has been found in over 85 countries in the world, the name of the disease is called Magnaporthe grisea (a fungal disease), It is capable of destroying food which is enough to feed more than 60 million people every year.

By image processing techniques and using neural networks , pest damage in pip fruits can be detected, here wavelets are used as a means by line detection, which was suggested by Woodford, Kasabov and Wearing in "Fruit Image Analysis using Wavelets"[7]. As the main research subjects, the leaf-roller, coding moth and apple leaf curling midge were chosen, because in orchards they were the predominant pests. Daubechies wavelets, derived from fast wavelets are used, to detect the main cause. In the first step of this two-step process, the elementary image is compared with the three color component deviation then in the second stage contains the calculated weighted version of the Euclidean distance between feature coefficient, which was chosen in the previous stage and the one with the querying image is calculated and the one with the least distance is arranged as identical image to conditions provided.

By CLASE (central lab of agricultural expert system) cucumber crop was handled from pest infection. Enhancement, segmentation, feature extraction and classification, this four main methods are usually used to investigate disorder from leaf image. Three different disorders such as leaf miner, powdery and downy are tested. By using these following methods errors have been highly reduced between systems.

Recognition of leaf disease[10] has been done by Prasad Babu and Srinivasa Rao using back propagation neural network. Only black propagation network is applicable to determine the species from the leaf, it is proven a fact.
to find back propagation algorithms they used Prewitt edge detection and thinning algorithm and use leaf token as input.to recognize various leaves with pest and rotten leaves due to insects Experimentation with large training sets is required or several reports predicted that diseases can be done as an advancement method.

To segment the agricultural land fields, using neural network, Remote sensing data was proposed [12].To apply a leaf recognition algorithm [12], high efficient recognition algorithm and easy to extract features are used. Probabilistic Neural Network approach is used for plant leaf recognition. This algorithm is 90% efficient on 32 kinds of plants.

In their report [13] Santanu and Jaya explained a software prototype system that can be used for disease recognition based on the diseased images of various rice plants. To detect infected parts of plants, image growing and image segmentation has been used.
Self-Organize Map (SOM) neural network has been used for classifying diseased rice images. For segmented leaf region, Otsu segmentation is used.
The results, calculation of the quotient of disease spot and leaf area plant diseases, are classified. Grape leaf disease is detected from colour imagery [14] by using a hybrid intelligent system. We used self-organizing maps and back propagation neural networks.
An advanced self-organizing feature map with genetic algorithms for optimization has been used to carry out segmentation of grape leaf disease and classification by using support vector machine and Gabor wavelet allowing to analyse and classify leaf disease has been used for filtering.
A method is mentioned in the report [17], to investigate plant diseases caused by spores. Histogram generation, the grey-level correction, image feature extraction and image sharpening has been carried out processing and analysed, the collared image is converted to a grey image.
Grey image is processed by edge enhancement by using a Median filter and canny edge algorithm. Morphological features like dilation, erosion, opening etc. have been operated after thresholding binary image captured.
Application of image processing and Support Vector Machine (SVM) for the early and accurate detection of rice diseases has been proposed by the report. The marks of rice diseases are classified and their texture and shape are extracted. Characteristic values of classification are set by selecting shape and colour texture of disease spot.
The SVM method was useful to classify rice bacterial leaf blight, rice sheath blight, and rice blast. The SVM was 97.2% efficient.
For clustering and classification of diseases that affect plant leaves, Otsu segmentation, K-means clustering and back propagation feed, forward neural network, are used.

### III. METHODOLOGY AND RESULT

#### A. Preprocessing

Techniques of image processing and classification are the base of the entire framework of our model. The digital image samples of different varieties of leaves from the environment with a digital camera. Every sample image was then perfectly analyzed and processed with the use of our proposed approach. For further classification of those samples for our required objective, extraction of useful features has been done.

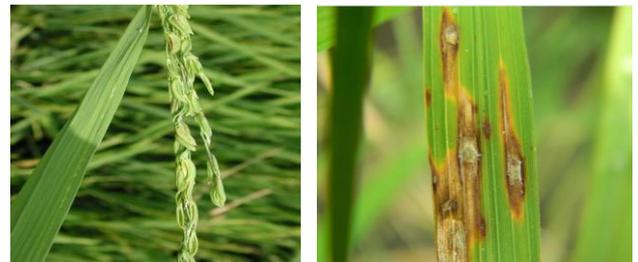

(a)        (b)
Fig 1: (a) normal rice leaf, (b) defective rice leaf

The main challenge in pre-processing section was to avoid noise due to the background which can severely affect the main result. So the background has been clipped using segmentation and Otsu's method of detecting grey threshold value.

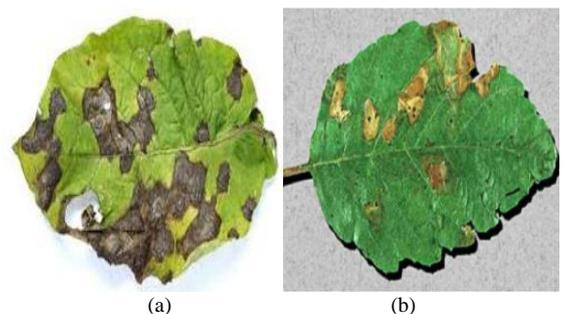

(a)        (b)

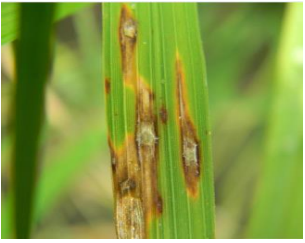
(c)
Fig. 2: Sample data of three defected leafs

### B. Histogram Equalization

As the preprocessing is done, before we segment and classify the leaf into the faulty and faultless regions, we enhanced the contrast and colour depth using the well-known histogram equalization technique. This technique widens the available range of intensities and is mapped to a more uniform distribution implemented with a remapping cumulative distribution function. For a histogram H(i) its remapping function will be

$$H'(i) = \sum_{0 \leq j < i} H(j)$$

After equalization, the new intensity values are given by,

$$equalized(x, y) = H'(src(x, y))$$

### C. Segmentation

As shown in fig. 3 the RGB images of all the leaf samples were studied in the initial time. The samples can be divided into two subclasses faultless and faulty. In the second step, the digital image has been converted to a binary image. This conversion is followed by the conversion to RGB image. These are the part of the color transformation. Image transformation is very important at this stage as it is the input by which K-means clustering is to be done for classification.

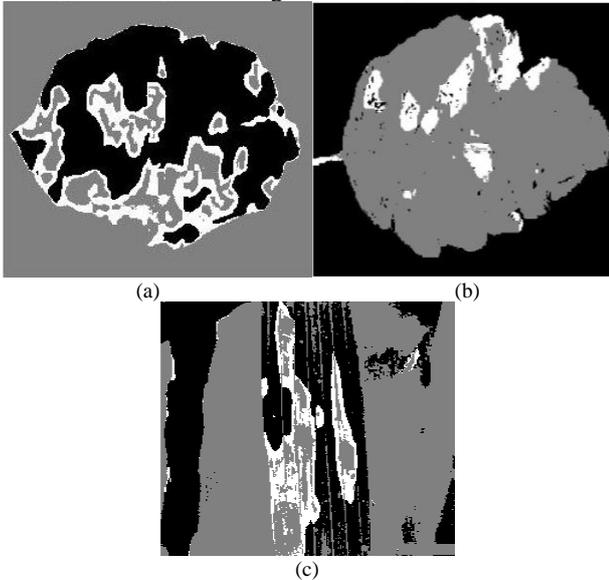
(a) (b)
(c)
Fig. 3: Colour transformation (RGB to Binary Image)

### D. Clustering

In computational intelligence, data clustering is a well-known procedure which is also used in pattern recognition. clustering means to join several objects into one or cluster in such a way that every object in a cluster is similar to other object and also different from the object present in other cluster. There are several ways of Clustering an image like, K-means Clustering, Fuzzy K-means Clustering, Clustering on the basis of density, Hierarchical Clustering. K-means is a centroid based algorithm, where k represents the number of clusters.

There are several advantages of K-means clustering like, if there are huge amount of variables then K-means clustering is faster than other clustering also K-means produces tighter cluster than any other clustering methods, especially if the clusters are globular, So here we are using k-means clustering as there are a huge amount of pixels to be clustered.

- Using k-means++ algorithm at first random two centers or pixels are chosen from the infected leaf. The centers represent the faulty and faultless regions of the leaf. It is done based on similar kind of featured weights. It is done to identify the infected cluster by a specific type of disease, of the sample leaf.
- Now the for all the pixels the nearest center is calculated and assigned to the corresponding centers
- At this stage the new two centers are calculated using the assigned pixels and the algorithm goes back to the previous step. This iterative process is followed till the centers stabilize.

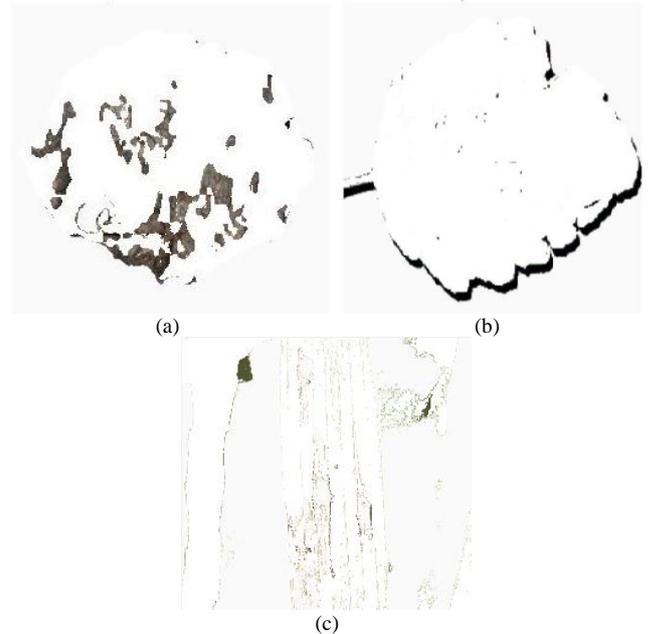
(a) (b)
(c)
Fig. 4: Clustered with threshold identification

The next is done in two steps, at first we have pointed out the green or approximately green pixels from the infected leaf . In order find accurately the varying threshold value which can minimize the infraclass variance of the threshold to find the black and white pixels, Otsu's method is applied

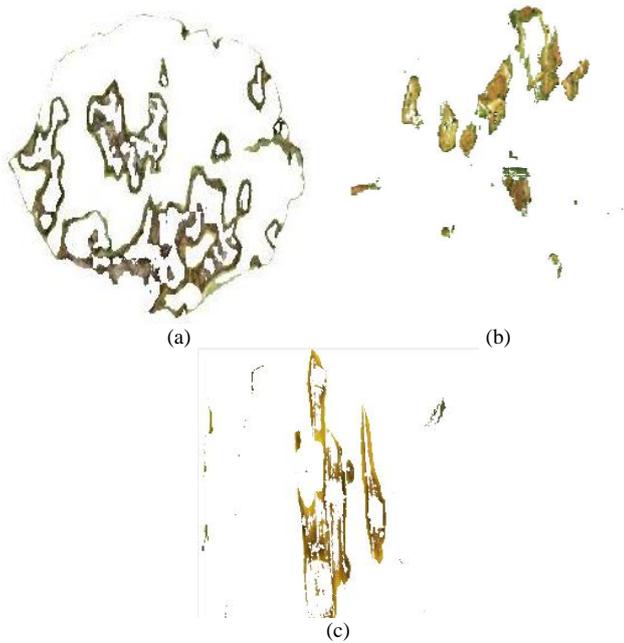

Fig. 5: Detected fault area of the leafs

At the second step, after removing all the green or apparently green pixels the first step is done in a loop till the optimized state is acquired. After getting the perfect image the result is quite impressive as the clearance is high than the previous image.

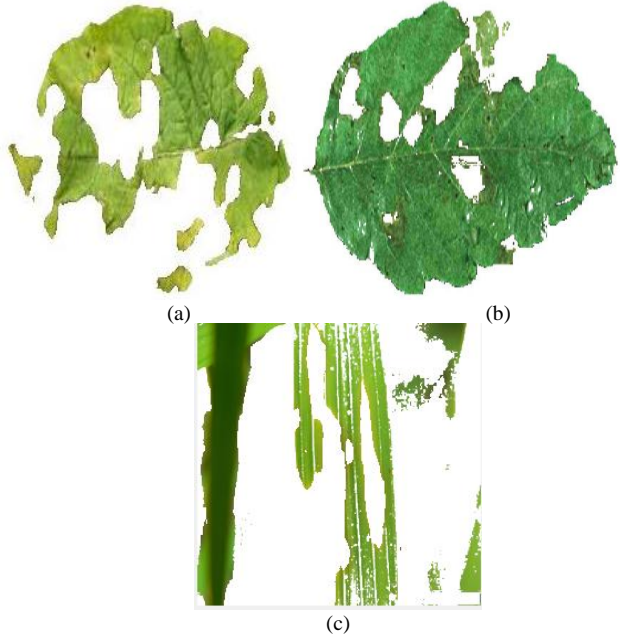

Fig. 6: Detected normal area of the leafs

After the previous step, the next step is mainly focused to find the ratio between the normal portion and the infected portion. This ratio is very important as from this ratio we can predict if this is curable or not. If the ratio lies lower than the expected ratio then the only way is to cut it off, or if it is higher than the expected ratio then it may be cured by fertilizer or medicine.

## IV. HISTOGRAM MATCHING & DETERMINATION OF RESULT ACCURACY

By far we have already segmented the leaf image and classified it into the faulty and faultless region. So the method that we have proposed successfully completed the challenge of detecting the leaf area affected by a disease. In this step, we cross-checked our results to ensure that the results found are accurate, which is done by checking and plotting the histograms shown in Fig. 7 below. The histograms of sample leaf, the faulty region, and the normal or faultless region are plotted in Fig. 7(a), Fig. 7(b) and Fig. 7(c) respectively.

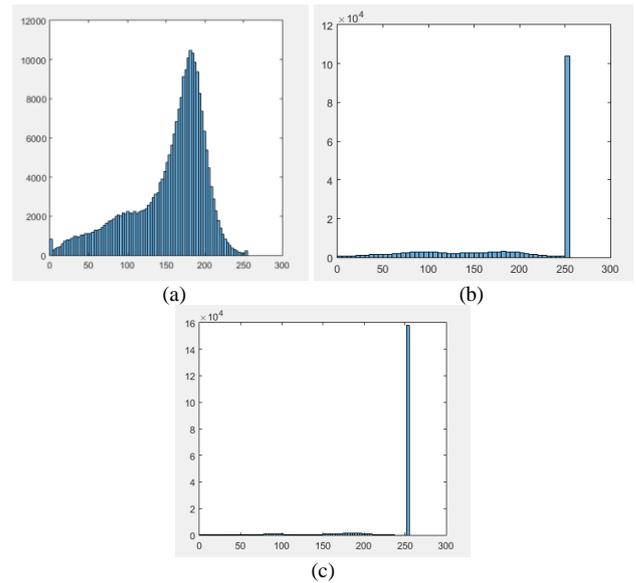

Fig. 7: Histogram plots of the sample leaf and its faulty and faultless area

## V. CONCLUSION

In this paper, we have addressed the major problem of different types of leaf diseases of the crops. With the help of image processing, we successfully segmented the leaf images and classified them into the faulty and normal region. Due to this segmentation, it becomes easier to detect if any leaf diseases affect the crops just with the help of an digital image. We used the unsupervised learning method by implementing the k-means clustering algorithm for segmentation of the faulty region in leaves which is way more accurate than the existing methods. This paper throws several new challenges to us, on which we are currently working. We are working to improve the accuracy further and determine a ratio of the normal and the faulty regions. This would be helpful to diagnose the disease and cure process to be effective and quick. It would also show if the disease not curable from the ratio and will help to set the further course of action.